\documentclass[runningheads]{breath_detector_cnn_transformer}

\usepackage[T1]{fontenc}
\usepackage{graphicx}
\usepackage{tikz}
\usepackage{array}
\usepackage{amsmath}
\usepackage{tabularx}
\usepackage{multirow}
\usepackage{url}
\usepackage{cite}
\usepackage{siunitx}
\usepackage[section]{placeins}
\usepackage[hidelinks]{hyperref}
\usepackage{makecell}
\usepackage{booktabs} 

\newcolumntype{Y}{>{\centering\arraybackslash}X}
\begin{document}

\title{
Acoustic Respiratory Phase Detection \\ on Mobile Devices 
}

\author{
Julian Szymański \and
Tomasz Sankowski \and
Piotr Sulewski \and
Iwo Czartowski \and 
Javier Medina Quero 
}


\author{Julian Szymański\inst{1} \and
Tomasz Sankowski\inst{1} \and
Piotr Sulewski\inst{1} \and
Iwo Czartowski\inst{1}
\and 
Javier Medina Quero\inst{2}
}

\authorrunning{J. Szymański et al.}

\institute{Faculty of Electronics, Telecommunications and Informatics,\\
Gdansk University of Technology, Poland\\
\email{\{julian.szymanski, tomasz.sankowski, piotr.sulewski, iwo.czartowski\}@pg.edu.pl}
\and
University of Granada, Granada, Spain\\
\email{javiermq@ugr.es}}

\maketitle

\begin{abstract}

Breathing monitoring is important for well-being, stress regulation, and guided breathing exercises, yet existing approaches often rely on contact-based sensors or controlled environments. This paper presents a non-invasive framework for respiratory phase detection using acoustic signals recorded on commodity smartphones. The proposed hybrid CNN--Transformer architecture combines local spectral feature extraction with long-range temporal modeling to capture the cyclic structure of breathing.

The model is evaluated using a Leave-One-Subject-Out (LOSO) protocol on the
Foundation Dataset and on a Independent unseen test set comprising over 1300
breathing cycles and 2 hours of continuous recordings. The proposed 3-class
formulation achieves 72.85\% frame-level accuracy and 96.6\% cycle-level
exhalation detection efficiency in the LOSO evaluation, while achieving
76.94\% frame-level accuracy, a Cohen's Kappa score of 0.65, and 99.2\%
cycle-level exhalation detection efficiency on the unseen test set.

Results suggest that accurate respiratory monitoring may be achievable under data-constrained conditions, indicating that appropriate temporal modeling can learn transferable respiratory representations and generalize to previously unseen participants. An end-to-end ONNX deployment pipeline enables privacy-preserving on-device inference with latency below 20 ms, supporting real-time monitoring on mobile and wearable devices.

\keywords{Respiratory Monitoring \and Breath Phase Detection \and CNN-Transformer \and Acoustic Sensing \and On-Device Inference}

\end{abstract}

\section{Introduction}

Breathing is a fundamental physiological process relevant to health monitoring, well-being, stress regulation, and guided breathing exercises. Continuous measurement of respiratory rate (RR) remains challenging, as conventional approaches often rely on manual counting or contact-based sensors such as spirometers and chest bands \cite{cretikos2008}. Acoustic monitoring using smartphones has emerged as a promising non-invasive alternative, with prior studies demonstrating that built-in microphones can reliably capture breathing sounds \cite{Nam_2016}. However, traditional signal-processing methods remain sensitive to noise and irregular breathing patterns.

Recent systematic reviews have highlighted the growing role of artificial
intelligence and audio analysis in respiratory assessment and disease detection,
demonstrating increasing interest in non-invasive acoustic monitoring
approaches \cite{kapetanidis2024audioreview}.

Advances in deep learning have improved audio classification performance. While recurrent architectures such as LSTMs have been widely applied to sequential tasks \cite{zaman2023audio}, their ability to capture long-range dependencies is limited. Transformers address this issue through self-attention mechanisms that model global temporal context in parallel \cite{vaswani2023attentionneed}. For respiratory monitoring, combining local spectral feature extraction with long-range temporal modeling is particularly attractive due to the inherently periodic nature of breathing.

Despite this progress, existing respiratory monitoring systems often depend on controlled environments, contact-based sensing, or subject-specific calibration, limiting their applicability in real-world mobile scenarios. Moreover, high frame-level classification accuracy does not necessarily translate into reliable respiratory rate estimation, where stable cycle-level detection is required.

To address these challenges, this work presents a hybrid CNN--Transformer architecture for real-time respiratory phase detection from acoustic signals recorded on commodity mobile devices. The proposed approach combines convolutional feature extraction with temporal attention mechanisms, enabling robust cycle-level respiratory monitoring in uncontrolled environments.

The main contributions of this work are as follows:

\begin{itemize}
\item \textbf{Hybrid CNN--Transformer Architecture:}
A CNN--Transformer model is proposed that combines local spectral feature extraction with long-range temporal modeling. Asymmetric pooling preserves temporal resolution, while a 3.5s sliding context window enables effective modeling of respiratory periodicity.

\item \textbf{Comparison of Classification Strategies:}
Binary (Exhale vs. Background) and ternary (Inhale, Exhale, Silence) formulations are evaluated. Results indicate that explicit modeling of all respiratory phases improves cycle-level stability, leading to more accurate RR estimation.

\item \textbf{Privacy-Preserving Mobile Deployment:}
An ONNX-based deployment pipeline is presented, enabling fully offline inference with latency below 20ms and ensuring that sensitive audio data never leaves the device.

\item \textbf{Dataset and Data-Efficient Learning:}
A publicly available respiratory audio dataset recorded in uncontrolled domestic environments using commodity hardware is introduced, supporting reproducible evaluation of acoustic respiratory monitoring.  Experiments indicate that useful respiratory phase detection and RR estimation performance can be obtained with a relatively small Foundation Dataset in the evaluated setting.

\end{itemize}

The presented system is intended as a practical solution for real-time breathing monitoring on mobile and wearable devices. 
The results suggest that effective temporal modeling can improve data efficiency when learning strongly periodic physiological signals, although further validation on larger datasets is required.
    
\section{Related Work}

Acoustic respiratory monitoring has progressed from manual feature engineering
to advanced deep learning. Recent reviews provide a comprehensive overview
of datasets, preprocessing techniques, and deep learning architectures used
for respiratory sound analysis \cite{huang2023lungreview}. Early smartphone-based efforts by Nam et al. \cite{Nam_2016} demonstrated the feasibility of using built-in microphones for RR estimation via envelope analysis. However, their reliance on magnitude-based processing made the system vulnerable to background noise and physiological variations, such as nasal congestion, often leading to frequency doubling errors.

To address these robustness issues, the focus of research transitioned toward Deep Learning architectures. Jácome et al. \cite{jacome2019cnn} utilized CNNs (Faster R-CNN) to detect breathing phases from lung sound spectrograms, attaining high sensitivity while remaining constrained to tracheal recordings. Subsequently, hybrid architectures combining CNNs with RNNs became the standard for capturing temporal dynamics. Acharya and Basu \cite{acharya2020deep} proposed a CNN-BiLSTM model for respiratory anomaly classification. Notwithstanding these advancements, the authors identified a critical bottleneck: high inter-patient variability, which limited generalized performance to approximately 66\%, necessitating complex patient-specific tuning. Furthermore, recurrent architectures process sequences step by step, which may limit their ability to exploit global context across a full breathing cycle.

Recent advancements have leveraged the Transformer architecture to overcome RNN limitations  \cite{vaswani2023attentionneed}. Literature reviews have also highlighted the growing role of attention-based architectures and transformer models in respiratory sound analysis and biomedical audio classification \cite{xu2024lungsoundsurvey}.

He et al. \cite{he2023multi} introduced the Multi-View Spectrogram Transformer, which processes time and frequency dimensions independently, outperforming traditional hybrid models on the ICBHI dataset. Similarly, lightweight attention mechanisms, such as the CNN-Temporal Self-Attention proposed by Fraihi et al. \cite{fraihi2025improving}, have demonstrated state-of-the-art results with reduced computational complexity. While other modalities like PPG \cite{samavati2024efficient} or IMU \cite{simone2024cardiorespiratory} also offer mobile respiratory estimation, acoustic analysis remains the most direct non-contact method.

This work focuses on respiratory phase detection in uncontrolled acoustic environments using a CNN--Transformer architecture with global temporal modeling. The resulting model is designed for efficient real-time deployment on mobile devices via ONNX.

\section{Method Description} \label{section/method}

The developed model is designed to process raw audio frames acquired from diverse input devices, including built-in smartphone microphones and external headsets. The primary objective is the precise classification of respiratory phases within each time frame. Adopting a supervised learning paradigm, the system implements a comprehensive pipeline encompassing robust signal preprocessing, dynamic data augmentation, and a hybrid neural network inference engine.

\subsection{Data Preprocessing and Augmentation}
The input is monophonic audio sampled at 44.1 kHz. Raw audio waveforms are generally suboptimal for direct classification due to high dimensionality and susceptibility to noise. Therefore, relevant features are extracted by converting the waveform into Mel-spectrograms. This representation was selected as it mimics the non-linear perception of pitch by the human ear, which is effective for biological signals.

\noindent The feature extraction process comprises three primary stages:
\begin{enumerate}
\item \textbf{Short-Time Fourier Transform (STFT):} The continuous signal is divided into frames using a window size of $n_{\text{fft}} = 2048$ samples with a hop length of 512 samples. This provides the necessary time-frequency resolution to detect short respiratory events.
\item \textbf{Mel Filtration:} The magnitude spectrum resulting from the STFT is projected onto 128 Mel bands ($n_{\text{mels}} = 128$).
\item \textbf{Logarithmic Scaling:} The amplitude is converted to a decibel scale. This compression is crucial for highlighting quiet breathing sounds (such as subtle inhalations) against the background noise floor.
\end{enumerate}

To prevent overfitting -- a common issue given the moderate size of the respiratory dataset -- and to improve generalization to different microphones and acoustic environments, a dynamic data augmentation pipeline was implemented. During training, the following transformations are applied stochastically to the input samples:
\begin{itemize}
\item \textbf{Gaussian Noise Injection:} Simulating lower-quality hardware or environmental background noise.
\item \textbf{Volume Scaling:} Multiplying the signal by a random factor (range [0.5,1.2]) to account for varying distances between the user and the microphone.
\item \textbf{Time Shift:} Cyclically shifting the audio along the time axis to prevent the model from memorizing the specific timestamps of breath occurrences.
\end{itemize}

\subsection{Hybrid CNN-Transformer Architecture}
\begin{figure}[ht]
\centering
\includegraphics[width=0.8\textwidth]{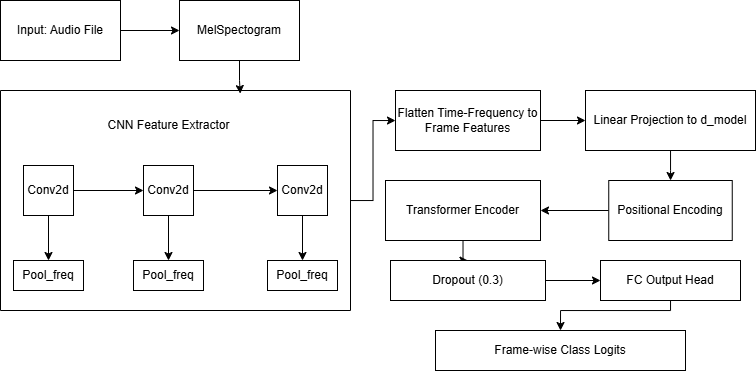}
\caption{The proposed hybrid architecture. Data flows through the CNN feature extractor, is augmented with positional encoding, and processed by the Transformer Encoder.}
\label{fig:arch}
\end{figure}

The core contribution of this work is the model architecture, visualized in Fig.~\ref{fig:arch}. It combines the strengths of Convolutional Neural Networks (CNNs) in extracting local spectral features with the capability of Transformers to model long-term temporal dependencies.

\noindent The architecture is composed of the following modules:

\subsubsection{CNN Feature Extractor.}
The initial stage consists of three convolutional blocks. Each block performs a 2D convolution, followed by Batch Normalization, ReLU activation, and Max Pooling.

A critical design decision was the implementation of asymmetric pooling with a kernel size of (2,1). While standard convolutional layers typically employ symmetric downsampling (e.g., 2×2), such an approach would lead to 'temporal blurring', effectively losing the fine-grained boundaries of breath sounds. By reducing dimensionality solely along the frequency axis -- compressing 128 Mel bands down to 16 abstract features -- the resolution of the time axis is strictly preserved. This temporal fidelity is essential for the sequence labeling task, as it ensures that the subsequent Transformer layers receive a high-density sequence.

\subsubsection{Positional Encoding and Transformer Encoder.}
The output of the CNN is a sequence of feature vectors. Since the standard Transformer architecture is invariant to the order of elements in a sequence, temporal information is injected using sinusoidal Positional Encoding. This adds sine and cosine functions of varying frequencies to the feature vectors, allowing the model to distinguish relative positions of frames.

The sequence is processed by a Transformer Encoder consisting of 6 layers with 8 attention heads, enabling long-range temporal modeling and contextual discrimination of respiratory sounds from background artifacts.

\subsubsection{Classification Head and Strategy Selection.}
Finally, a fully connected layer projects the high-dimensional output of the Transformer to class probabilities for each time frame. Two configuration strategies were tested:
\begin{enumerate}
\item \textbf{3-Class Model (Ternary):} Trained to distinguish between Inhalation, Exhalation, and Silence.
\item \textbf{2-Class Model (Binary):} Optimized to detect Exhalation versus Background (aggregating Inhalation and Silence).
\end{enumerate}

\subsection{Sliding Window Mechanism for Real-Time Inference}

A critical challenge in processing live audio streams is the fragmentation of data into short packets (typically \SI{200}{ms}), which lack sufficient temporal dynamics for accurate phase recognition. Direct classification of such short segments results in poor model performance due to the absence of broader context.

To resolve this, a sliding window buffer with a fixed length of \num{154350} samples (approximately \SI{3.5}{s} at \SI{44.1}{kHz}) was implemented. This mechanism operates as a First-In-First-Out (FIFO) queue: new audio samples are appended to the buffer's end, while the oldest samples are discarded from the beginning. This approach ensures that every inference step utilizes a comprehensive temporal context. The window length was chosen to cover most of a typical resting breathing cycle, so that the model observes neighbouring respiratory phases when classifying the current one.

\subsection{Baseline Models for Comparative Analysis}
To evaluate the necessity of the hybrid approach, two simpler baseline architectures were implemented:

\begin{itemize}
\item \textbf{Feed-Forward Neural Network (FFNN):} This model processes each Mel-spectrogram frame as an independent 128-dimensional feature vector. The architecture consists of a multi-layer perceptron with three hidden layers (256→256→128 units), each followed by Batch Normalization, ReLU activation, and Dropout (p=0.3).
\item \textbf{Convolutional Neural Network (CNN):} This model employs a four-block 2D convolutional architecture with (3,1) kernels that operate exclusively along the frequency axis. Each block consists of Conv2D, Batch Normalization, ReLU, and MaxPooling (applied only vertically). This baseline isolates the contribution of spectral feature learning from temporal modeling.
\end{itemize}

\section{Experiments and Results}

To evaluate the performance of the proposed architecture and verify the impact of temporal context modeling, a series of experiments was conducted using four model variants: Feed-Forward (FF), Convolutional (CNN), and two Hybrid CNN-Transformer models (3-class and 2-class).

\subsection{Evaluation Methodology and Datasets}

The experiments were conducted on a  dataset of ``nose-only'' breathing recordings collected in uncontrolled domestic environments. The evaluation follows a dual-track protocol to rigorously assess both baseline algorithmic generalization and independent real-world applicability:

\begin{itemize}
    \item \textbf{Track 1: Foundation Dataset (LOSO Validation):} 
    Recordings from three male researchers (ages 21--23) containing 822 annotated breathing cycles. To strictly evaluate generalization without data leakage, this set was evaluated using a Leave-One-Subject-Out (LOSO) cross-validation protocol. In each fold, the model was trained on two subjects and evaluated on the third unseen subject.
    \item \textbf{Track 2: Independent Unseen Dataset:} 120 continuous \SI{60}{\second} recordings (2 hours in total) from 13 previously unseen participants (ages 14--52), containing over 1300 breathing cycles. The model evaluated here was trained on the entire Foundation Dataset.
\end{itemize}

This dual-track setup provides complementary assessments of cross-subject
robustness under limited-data conditions. The LOSO protocol evaluates
generalization to an unseen subject when training on two participants,
while the Independent Unseen evaluation assesses performance on a diverse
cohort of 13 participants after training on the complete Foundation Dataset.
All participants gave informed consent prior to recording; for participants
under 18, consent was additionally obtained from a parent or legal guardian.

\subsection{Leave-One-Subject-Out (LOSO) Analysis: The Frame vs. Cycle Paradox}

The LOSO evaluation on the Foundation Dataset reveals a critical paradox between frame-level accuracy and practical cycle-level detection (Table~\ref{tab:loso_results}).

\begin{table}[ht]
\caption{LOSO cross-validation results comparing frame-level accuracy and cycle-level exhalation detection efficiency.}
\label{tab:loso_results}
\centering
\begin{tabular}{lcccc}
\toprule
\multirow{2}{*}{\textbf{Subject}} & \multicolumn{2}{c}{\textbf{Frame Accuracy}} & \multicolumn{2}{c}{\textbf{Cycle Detection (Exhales Missed/Total)}} \\
\cmidrule(lr){2-3} \cmidrule(lr){4-5}
& \textbf{2-Class} & \textbf{3-Class} & \textbf{2-Class} & \textbf{3-Class} \\ 
\midrule
S1     & 88.37\% & 77.63\% & 0 / 186   & 0 / 186 \\
S2   & 81.77\% & 77.08\% & 80 / 309  & 19 / 309 \\
S3  & 81.80\% & 63.84\% & 59 / 327  & 9 / 327 \\
\midrule
\textbf{Average/Total} & \textbf{83.98\%} & \textbf{72.85\%} & \textbf{139 / 822} & \textbf{28 / 822} \\
\bottomrule
\end{tabular}
\end{table}

On the surface, the 2-class (binary) model appears superior, achieving an average frame-level accuracy of 83.98\% compared to 72.85\% for the 3-class (ternary) model. However, this metric is skewed by class imbalance, as the binary model merges inhalation and silence into a dominant ``background'' class.

When evaluated on the primary objective --cycle-level breath counting -- the 3-class architecture outperforms the binary approach. Across all LOSO folds, the 3-class model missed only 28 out of 822 exhalations (96.6\% detection efficiency), whereas the 2-class model missed 139 events.


This discrepancy is best illustrated by the results for subject ``S3''. While the 3-class model struggled with frame-level accuracy for this subject, achieving 63.84\%, it demonstrated remarkable resilience in cycle tracking, missing only 9 out of 327 exhalations. 
This suggests that explicitly modeling the complete respiratory sequence (Silence $\rightarrow$ Inhale $\rightarrow$ Exhale $\rightarrow$ Silence) may act as a structural regularizer. We hypothesize that even when the inhalation acoustic signature of a new user is distorted or faint, the 3.5\,s context window allows the model to exploit the surrounding respiratory structure, anchoring exhalation detection. Given that the LOSO evaluation includes only three subjects, this interpretation should be treated as preliminary.

\subsection{Independent Generalization and Statistical Analysis}

Building upon the generalization performance observed in the LOSO protocol, the final models were evaluated on the Independent Unseen Dataset (Track 2) containing 13 new participants. 
To quantify the uncertainty of the observed performance differences,
a bootstrap resampling method ($N=1000$ iterations) was employed
to estimate 95\% confidence intervals (CI) of frame-level accuracy.

\begin{table}[ht]
\caption{Frame-level performance on the Independent Unseen Dataset (13 subjects, 2 hours). Accuracy is given as mean $\pm$ standard deviation over $N=1000$ bootstrap samples, together with the 95\% confidence interval.}
\label{tab:stats_unseen}
\centering
\begin{tabularx}{\textwidth}{|l|Y|Y|Y|}
\hline
\textbf{Architecture} & \textbf{Acc. (Unseen)} & \textbf{95\% CI} & \textbf{Kappa} \\ \hline
Feed-Forward (FF) & 58.08\% $\pm$ 0.3\% & [57.6, 58.6] & 0.374 \\ \hline
Pure CNN & 60.53\% $\pm$ 0.3\% & [60.0, 61.1] & 0.390 \\ \hline
\textbf{3-Class Transformer} & 76.94\% $\pm$ 0.2\% & [76.5, 77.4] & \textbf{0.651} \\ \hline
2-Class Transformer & \textbf{83.10\% $\pm$ 0.2\%} & \textbf{[82.7, 83.5]} & 0.630 \\ \hline
\end{tabularx}
\end{table}

As demonstrated in Table~\ref{tab:stats_unseen}, context-free models (FF, CNN) experience severe performance degradation on this expanded cohort, dropping to approximately 60\% accuracy. 
Lacking temporal context, these models appear to be more sensitive to subject-specific acoustic characteristics.

Conversely, the Transformer-based architectures demonstrate robustness. Consistent with the LOSO findings, the 2-class model's higher raw accuracy (83.10\%) masks underlying instabilities. 
Cohen's Kappa, which accounts for chance agreement under class imbalance, is 0.651 for the 3-class and 0.630 for the 2-class Transformer. Since Kappa values computed over different numbers of classes are not directly comparable, we treat this only as supporting evidence and base the comparison of the two formulations primarily on the cycle-level and RR metrics in Sect.~\ref{sec:rr}. Part of the remaining frame-level error is likely related to uncertainty of manual phase-boundary annotation, which was not quantified in this study.

\subsection{Comparison with State-of-the-Art Approaches}

To contextualize the performance of the proposed model, a qualitative comparison with recent state-of-the-art (SOTA) methods for acoustic respiratory analysis is provided. Direct quantitative comparison is challenging due to differences in datasets, recording conditions, and evaluation protocols. Most existing approaches are evaluated on clinical datasets such as ICBHI, which focus on pathological sound classification (e.g., wheezing, crackles) rather than continuous respiratory phase detection in unconstrained environments.

For instance, CNN-RNN hybrid models, such as the CNN-BiLSTM approach proposed by Acharya and Basu~\cite{acharya2020deep}, report generalization performance of approximately 66\% in cross-patient settings, highlighting the difficulty of modeling inter-subject variability. More recent Transformer-based architectures, such as the Multi-View Spectrogram Transformer by He et al.~\cite{he2023multi}, achieve strong performance on controlled benchmark datasets, but are not explicitly designed for real-time respiratory phase tracking on mobile devices.

In contrast, the proposed hybrid CNN-Transformer model is evaluated in a more challenging real-world scenario, involving non-contact acoustic recordings in uncontrolled domestic environments and explicit testing on unseen subjects. 
Under these conditions, the model achieves 76.94\% frame-level accuracy and 99.2\% cycle-level detection efficiency on unseen data, indicating promising performance on the evaluated unseen subjects despite limited training data.

Furthermore, unlike many existing methods that focus solely on classification accuracy, this work emphasizes cycle-level consistency and respiratory rate estimation, which are directly relevant for real-time breathing monitoring and feedback applications. This distinction makes direct numerical comparison difficult but indicates potential advantages of the proposed approach in real-world mobile and wearable deployment. For clarity, a qualitative summary of key differences is presented in Table~\ref{tab:sota_comparison}.

\begin{table}[ht]
\caption{Qualitative comparison with representative approaches.}
\label{tab:sota_comparison}
\centering
\begin{tabular}{|l|l|l|c|c|}
\hline
Method & Dataset & Task & Real-time & Non-contact \\ \hline
CNN-BiLSTM~\cite{acharya2020deep} & Controlled & Classification & No & No \\ \hline
\makecell{Spectrogram\\Transformer~\cite{he2023multi}} & ICBHI & \makecell{Pathology\\detection} & No & No \\ \hline
\textbf{Proposed} & Uncontrolled & Phase + RR & Yes & Yes \\ \hline
\end{tabular}
\end{table}

As shown in Table~\ref{tab:sota_comparison}, the proposed method combines real-time capability and non-contact sensing in uncontrolled environments.

\subsection{Error Typology: Phase Fragmentation vs. Boundary Ambiguity}
Qualitative analysis reveals a fundamental distinction in temporal consistency. The FF and CNN models exhibit high sensitivity to transient acoustic artifacts, such as background noise, short clicks, or the natural volume attenuation occurring at the end of a respiratory phase. Lacking a global context window, these baselines frequently misinterpret such events, triggering premature or spurious state transitions that fragment a single respiratory phase into multiple unstable predictions. 

In contrast, the Hybrid CNN-Transformer architecture demonstrates higher classification confidence and stability. By leveraging its 3.5s global context, the Transformer maintains the current state until the broader rhythmic pattern confirms a genuine transition. Consequently, its inaccuracies are largely confined to boundary ambiguity -- slight shifts in phase onset or offset by approximately $0.2$-$0.4s$ -- which preserves the integrity of each respiratory cycle and ensures reliable tracking required for real-time breathing monitoring.

\begin{figure}[!t]
\centering
\includegraphics[width=0.8\textwidth, trim=0 0 0 1.2cm, clip]{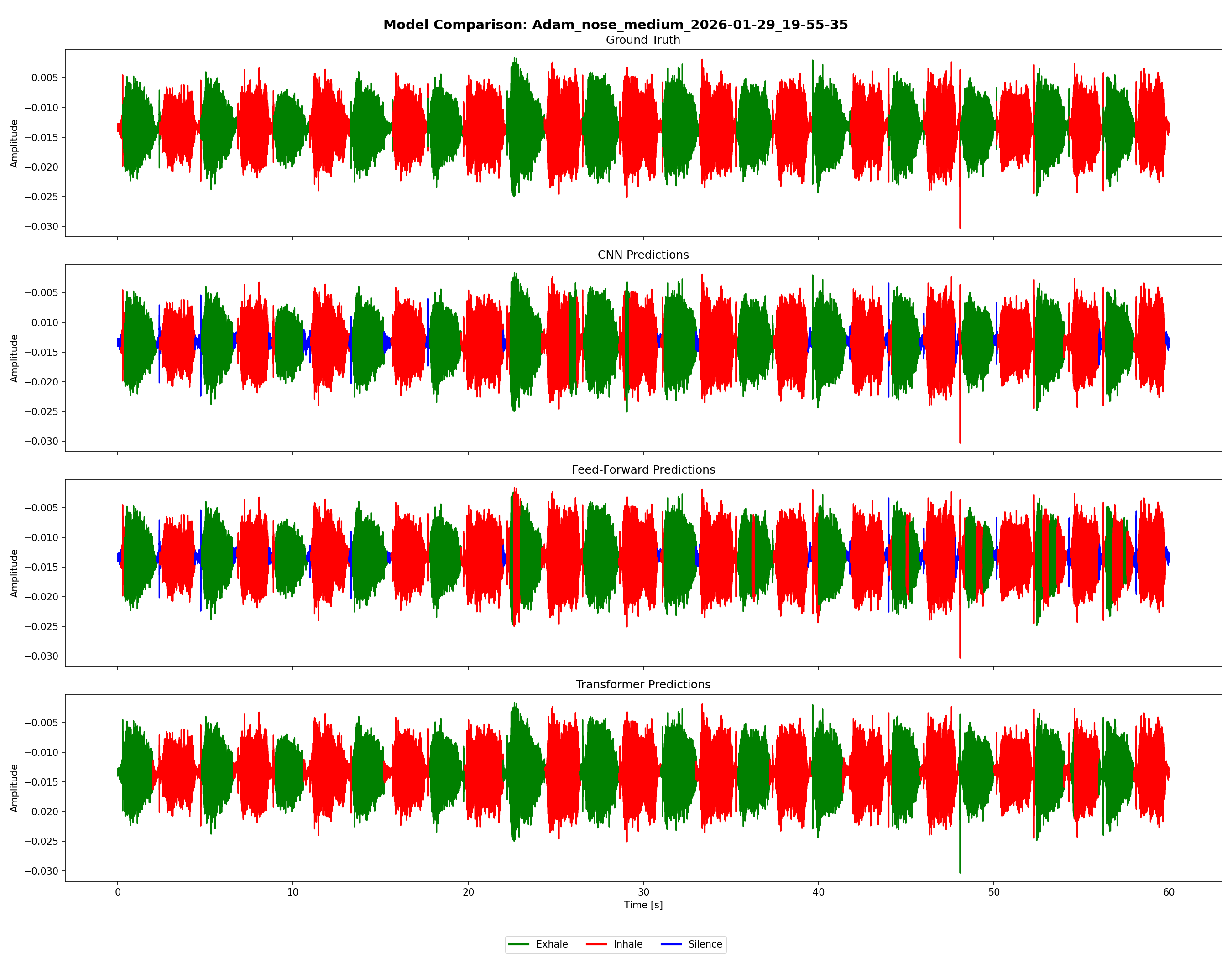} 
\caption{Inference pattern comparison for a subject unseen during training. The Hybrid Transformer (bottom) maintains high temporal continuity, while context-free models (FF, CNN) exhibit internal phase fragmentation.}
\label{fig:hyper}
\end{figure}

\begin{table}[!htbp]
\caption{Cycle-level exhalation detection efficiency and RR estimation on the Independent Unseen Dataset.}
\label{tab:rr_accuracy}
\centering
\begin{tabular}{|l|c|c|c|}
\hline
\textbf{Architecture} & \textbf{Detection Efficiency} & \textbf{MAE (BPM)} & \textbf{\% within $\pm$2 BPM} \\ \hline
2-Class Transformer & 92.8\% (1216/1310) & 1.38 & 83.3\% \\ \hline
\textbf{3-Class Transformer} & \textbf{99.2\% (1300/1310)} & \textbf{0.93} & \textbf{90.0\%} \\ \hline
\end{tabular}
\end{table}

\subsection{Cycle-Level Efficiency and Respiratory Rate Estimation}\label{sec:rr}

The practical utility of the system depends on the accurate estimation of the Respiratory Rate (RR) for real-time breathing feedback. Due to phase fragmentation (Sect.~4.5, Fig.~\ref{fig:hyper}), the baseline FF and CNN models split single respiratory phases into multiple segments, which makes cycle counting unreliable; cycle-level evaluation is therefore reported for the Transformer-based models only.

Following the high cycle-level detection performance observed in
the LOSO analysis, the respiratory rate estimation performance
on the Independent Unseen Dataset (13 subjects) is summarized in
Table~\ref{tab:rr_accuracy}.

The proposed 3-class Transformer model achieves strong cycle-level
performance on the Independent Unseen Dataset, with a cycle-level
exhalation detection efficiency of 99.2\% and a mean absolute error
(MAE) of 0.93 BPM. Moreover, 90\% of predictions fall within a
$\pm$2 BPM error margin. These results indicate that the model can
maintain reliable respiratory cycle tracking and RR estimation on
previously unseen participants.

\section{Practical Deployment and Dataset Considerations}

The value of this work extends beyond the proposed architecture, encompassing both the creation of a specialized respiratory dataset and the demonstration of privacy-preserving real-time deployment.

A primary contribution of this research is the development of an annotated respiratory audio dataset recorded in uncontrolled domestic environments. Unlike publicly available datasets such as ICBHI, which primarily focus on respiratory pathologies, the presented dataset captures ``nose-only'' breathing for continuous respiratory monitoring using commodity mobile devices. By incorporating variations in recording distance, ambient noise, and microphone quality, the dataset reflects practical conditions encountered in non-contact smartphone sensing applications.

Furthermore, the proposed CNN--Transformer architecture is not inherently limited to respiratory acoustics. By combining local feature extraction with long-range temporal modeling, the framework may be applicable to other physiological time-series and audio analysis tasks where both short-term patterns and temporal context are important. However, evaluation of such applications remains outside the scope of the present work.

The deployment pipeline uses ONNX Runtime for fully local inference, ensuring that audio remains on the user's device. The exported model achieves inference latency below 20\,ms and can be integrated into mobile and wearable applications without retraining.

\section{Discussion and Future Work}

The results presented in this work indicate that the proposed hybrid CNN--Transformer architecture is effective
 for acoustic respiratory monitoring on mobile devices. By combining local spectral feature extraction with long-range temporal modeling, the system achieves high cycle-level stability and accurate respiratory rate estimation. The obtained MAE of 0.93 BPM and 99.2\% exhalation detection efficiency
on previously unseen participants support the feasibility of the proposed
approach for real-time breathing monitoring in mobile and wearable applications.

A key observation is that the Transformer-based models substantially outperform the context-free baselines in terms of temporal consistency. By leveraging a 3.5s context window and explicitly modeling the complete respiratory cycle (Inhale--Exhale--Silence), the proposed architecture reduces phase fragmentation and produces more reliable cycle-level predictions. These findings suggest that temporal context modeling is particularly well suited for strongly periodic physiological signals such as breathing.

An important objective of this study was to investigate data-efficient learning
under realistic mobile sensing conditions. For the Independent Unseen
evaluation, the model was trained on the complete Foundation Dataset collected
from three participants and evaluated on 13 entirely unseen participants,
covering a broader age range (14--52 years) and more than 1300 breathing
cycles. In the LOSO evaluation, each model was instead trained on two
participants and evaluated on the third. These results provide an initial
indication of cross-subject transfer under limited training data. At the same
time, the demographic composition of the Foundation Dataset limits the
assessment of population-level generalization, as it contains only three male
participants aged 21--23. Future work should therefore evaluate the model on
larger and more diverse cohorts and report subgroup-level performance.

Overall, the presented results suggest that appropriate temporal modeling can substantially improve respiratory phase detection and respiratory rate estimation while reducing the dependence on large annotated datasets. The proposed architecture therefore represents a promising direction for privacy-preserving and accessible breathing-monitoring systems operating directly on consumer devices.

Several limitations should be considered when interpreting the results. First, the Foundation Dataset contains only three participants, which restricts the assessment of generalization. Second, respiratory rate ground truth was derived from manual annotation of the same audio recordings rather than from an independent reference sensor, and annotation uncertainty at phase boundaries was not quantified. Third, cycle-level evaluation focuses on missed exhalations; a detailed analysis of spurious detections is left for future work. Fourth, all recordings contain nose-only breathing. Finally, the study does not include an ablation of the context length or a comparison with recurrent temporal models using the same context window, so the observed gains can be attributed to temporal context modeling in general rather than specifically to self-attention. 

Future work should explore several directions to further improve the applicability and robustness of the proposed approach. The architecture may be extended to other respiratory-related acoustic events, such as coughing or snoring, where temporal context may also play an important role. Further studies should evaluate the system on larger and more diverse populations, including different age groups, physiological characteristics, and recording conditions, in order to better assess its generalization capabilities. In addition, incorporating pathological respiratory sounds, such as wheezing or stridor, could extend the system toward preliminary clinical screening. Combining acoustic sensing with smartphone IMU data may further improve robustness and accuracy, particularly under challenging recording conditions.

\section*{Data Availability}
To ensure reproducibility, the complete framework -- including all model architectures, evaluation scripts, data and a sample Flutter application utilizing the ONNX model for real-time classification on mobile devices -- is available at: \url{https://github.com/peterprospl12/breathing-classification-v2}.

\section*{Acknowledgments}
The work was supported partially by 
Funds of Faculty of Electronics, Telecommunications and Informatics Gdańsk University of Technology.

\bibliography{bibliography}

\end{document}